\documentclass{article}

\usepackage{microtype}
\usepackage{graphicx}
\usepackage{subcaption}
\usepackage{pgfplots}
\usepackage{booktabs} 
\usepackage[title]{appendix}

\usepackage{hyperref}
\usepackage{icml2019}
\usepackage{array}
\newcolumntype{P}[1]{>{\centering\arraybackslash}p{#1}}

\begin{document}

\twocolumn[
\icmltitle{Reducing Risk of Model Inversion Using Privacy-Guided Training}


\icmlsetsymbol{equal}{*}

\begin{icmlauthorlist}
\icmlauthor{Abigail Goldsteen}{ibm}
\icmlauthor{Gilad Ezov}{ibm}
\icmlauthor{Ariel Farkash}{ibm}
\end{icmlauthorlist}

\icmlaffiliation{ibm}{Data Security and Privacy, IBM Research Haifa, Haifa, Israel}

\icmlcorrespondingauthor{Abigail Goldsteen}{abigailt@il.ibm.com}

\icmlkeywords{Machine Learning, Privacy, Risk, Model inversion, Attribute inference, Feature influence}

\vskip 0.3in
]



\printAffiliationsAndNotice{\icmlEqualContribution} 

\begin{abstract}
Machine learning models often pose a threat to the privacy of individuals whose data is part of the training set. Several recent attacks have been able to infer sensitive information from trained models, including \emph{model inversion} or \emph{attribute inference} attacks. These attacks are able to reveal the values of certain sensitive features of individuals who participated in training the model. 
It has also been shown that several factors can contribute to an increased risk of model inversion, including feature influence. We observe that not all features necessarily share the same level of privacy or sensitivity. In many cases, certain features used to train a model are considered especially sensitive and therefore propitious candidates for inversion.

We present a solution for countering model inversion attacks in tree-based models, by reducing the influence of sensitive features in these models. 
This is an avenue that has not yet been thoroughly investigated, with only very nascent previous attempts at using this as a countermeasure against attribute inference. 
 Our work shows that, in many cases, it is possible to train a model in different ways, resulting in different influence levels of the various features, without necessarily harming the model's accuracy. We are able to utilize this fact to train models in a manner that reduces the model’s reliance on the most sensitive features, while increasing the importance of less sensitive features. Our evaluation confirms that training models in this manner reduces the risk of inference for those features, as demonstrated through several black-box and white-box attacks. 
\end{abstract}


\section{Introduction}
\label{sec:intro}
Machine learning models have been shown to pose a threat to the privacy of individuals whose data is part of the training set. Over the past few years, several attacks have been able to infer sensitive information from trained models. Examples include: \emph{membership inference attacks}, where one can deduce whether a specific individual was part of the training set or not; and \emph{model inversion attacks}, also called \emph{attribute inference attacks}, where certain sensitive features can be inferred about individuals whose data was used to train a model. 

In model inversion (MI) attacks, a trained ML model is used, along with additional information about an individual, to infer the value of one or more sensitive attributes used in the training of the model. These attacks may be performed either in black-box manner, meaning that only the output of the model on a particular input is known, or in a white-box manner where internal parameters of the model are also known. The additional information may include the values of other features used in the model training process. Typically, such attacks are employed against individuals whose data is part of the training set, since their success rate is higher in this setting.

This has led some experts to conclude that machine learning models themselves can be considered personal information under different data protection regulations such as the EU General Data Protection Regulation (GDPR)\footnote{\url{https://ec.europa.eu/info/law/law-topic/data-protection/data-protection-eu\_en}} and the California Consumer Protection Act (CCPA)\footnote{\url{https://leginfo.legislature.ca.gov/faces/billTextClient.xhtml?bill\_id=201720180AB375}} \cite{Veale18}. This creates an even greater incentive for organizations to protect machine learning models from such attacks, or even train models in a way that prevents them from being susceptible to these attacks to begin with.

Many methods have been proposed to anonymize or de-identify the training data of ML models, mostly focused on preventing membership inference attacks. Far less attention has been dedicated to attribute inference, with only a few, initial investigations of the subject \cite{Zhao19, Alves19}. Fredrikson et al.  \cite{Fredrikson15} proposed a simple countermeasure to their own attack against decision trees, by restricting the levels at which sensitive attributes may appear in the tree.
More recent works have begun attempting to analyze and quantify the risk of model inversion, as well as understand the risk factors that contribute to the success of such attacks. Some of these works have identified \emph{feature influence} as one of the major contributing factors to attribute inference.

Building on these discoveries regarding the effect of a feature's influence on the ability to invert it, and on the initial attempts at countermeasures introduced for decision tree models, we propose a new \emph{privacy-guided} approach to training tree-based models. Our stretegy is based on guiding the model training process to rely less on those features defined as most sensitive, by reducing the amount of times they are used in the model. Reducing how often the feature is used by the model to make decisions, will make it more difficult for attackers to succeed in extracting or inferring information about it from the model.

Not all features necessarily have the same level of privacy or sensitivity. In many cases, certain features used to train a model are considered especially sensitive and therefore propitious candidates for inversion. We utilize this fact to train models in a manner that reduces the importance of the most sensitive features, in turn increasing the importance of the less sensitive features.

We propose and evaluate three slightly different methods to achieve this reduction in feature importance. We implemented and evaluated these methods in decision tree, random forest, and AdaBoost models. We first show that each of these methods does have some effect on the importance of the feature, although some methods work better than others in certain combinations of datasets, features, and models. We then evaluate their effect on the success rate of model inversion attacks on decision trees. Our results show that it is indeed possible to reduce the success rate of such attacks by reducing the importance of the attacked feature in the model. In addition we demonstrate that, in many cases, reducing the importance of one or more features has a negligent effect on the model's accuracy, and may even come at no cost at all in terms of model accuracy. 
 
 
 Tha main contributions of this work are: (1) Three complimentary methods for reducing the importance of sensitive features in tree-based models; (2) Empirical validation that reducing the importance of certain features can be done without affecting the accuracy of the model; and (3) Additional corroboration that a feature's importance plays a pivotal role in the ability to invert it through model inversion attacks, and that reducing a feature's importance can help protect it from such attacks.
 
In the remainder of the paper we present the most relevant background work in Section \ref{sec:back}, describe our proposed method in Section \ref{sec:priv}, and present our experimental results in Section \ref{sec:eval}. We survey additional related work in Section \ref{sec:rel} and finally conclude in Section \ref{sec:con}.

\section{Background}
\label{sec:back}

One of the first successful model inversion attacks was realized in the domain of pharmacogenetics. Fredrikson et al. \cite{Fredrikson14} demonstrated that, given a machine learning model for Warfarin dosing and some demographic information about a patient, it was possible to predict the patient’s genetic markers. In this case, the model was a linear regression model that predicted a real-valued suggested initial dose of the drug Warfarin. The attack was a black-box attack, where the attacker had access only to the output of the model for a given input. 

Fredrikson, Jha, and Ristenpart \cite{Fredrikson15} later developed a new class of model inversion attack that exploits the confidence values revealed along with the predictions. They employed both black and white-box attacks, and showed their effectiveness in decision-tree models and neural networks. They also initiated the experimental exploration of natural countermeasures, to examine whether the level in the tree at which the sensitive feature occurs can affect the accuracy of the attack. They developed a privacy-aware decision tree training algorithm that is a simple variant of Classification And Regression Tree (CART) learning. It takes a parameter $l$ that specifies the priority at which the sensitive feature is considered: the feature is only considered for splitting after $l-1$ other features have already been selected, and removed from consideration afterwards. This is referred to by the authors as a preliminary investigation designed to guide future countermeasure design; it was evaluated on a single feature and dataset. During this experiment, they discovered that when the feature appears near the top or bottom of the tree, the attack fails with greater probability than otherwise. They also found that when the feature is placed at the top of the tree, classification accuracy is maximized while inversion accuracy is greatly reduced. 

More recently, a generative adversarial network (GAN) based black-box model inversion attack has been shown to be effective against deep-learning models such as convolutional neural networks \cite{Aivodji19}. This work was able to extract recognizable features from both image recognition and skin classification models.

In 2016, a body of work began attempting to analyze and quantify the risk of model inversion, as well as understand the risk factors that contribute to the success of such attacks. 
Wu et al. \cite{Wu16} started by presenting a game-based methodology to formally study model inversion attacks. They were the first to talk about \emph{influence} as a characteristic of MI attacks, and suggested that adding noise may help to counter this effect. Their work provides a methodology for both black-box and white-box attacks. Their analysis is not limited to a certain type of model, but does make certain assumptions about the model and features: it assumes a binary classification task where all features are binary. The authors state that this methodology can also be extended to binary classification over generalized but finite domains. 

Around the same time, Papernot et al. \cite{Papernot16} created a comprehensive threat model for ML, categorizing and taxonomizing attacks and defenses within an adverserial framework. To solve the issue of learning and inferring with privacy, they suggested using differential privacy techniques, by injecting random noise into the data, the cost function of the learning algorithm, or the learned parameters. However, they themselves note that learning models with differential privacy guarantees is difficult because the sensitivity of the models is unknown for most interesting ML techniques. 

Yeom et al. \cite{Yeom18} continued to examine the effect of overfitting and influence on the ability of an attacker to learn information about the training data from machine learning models. Using both formal and empirical analyses, they illustrated a clear relationship between these factors and the privacy risk that arises in several popular machine learning algorithms. More specifically, they showed that both overfitting and influence play an important role in attribute inference, and found that the risk to individuals in the training data is greatest when these two factors are in balance. Influence in this case is defined as the magnitude of change to the outcome as a result of changing the sensitive feature’s value.

\section{Privacy-Guided Training of Models}
\label{sec:priv}
Our approach is based on the assumption that reducing the importance of a feature in the training of a model will in turn reduce the risk of it being inverted by MI attacks. We intuit that if a feature has little to no influence on the outcome of the model, it will be difficult to use this outcome to infer any meaningful information about the feature. If, on the other hand, a feature is highly correlated with the outcome of the model, it will be easier to deduce something about its content. In the extreme example of a single feature having 100\% importance, the outcome has a one-to-one correlation with the feature. 

In previous work \cite{Wu16, Yeom18}, the authors found feature influence to be an important factor in incurring privacy risk, alongside overfitting. In the context of Boolean functions, Wu et al. \cite{Wu16} defined influence as the probability that changing the sensitive feature's value will cause a change in the model's outcome. In the context of linear regression models, Yeom et al. \cite{Yeom18} considered an analogous definition of influence that characterizes the magnitude of change to the outcome. In linear models, this corresponds to the absolute value of the normalized coefficient of the sensitive feature.

In our setting, we use the feature's \emph{importance}, which is a model-dependent definition. In decision trees (and similarly in random forests and other tree-based models), importance is usually calculated as the decrease in node impurity weighted by the probability of reaching the node, summed over all nodes that split on the given feature. We used the following equation for calculating importance values:
\begin{equation}
Imp(f)=\sum_{i} (S(i) - E(l_i) - E(r_i)) \cdot \frac{n_i}{N}
\end{equation}
Where $i$ enumerates over all nodes that split on feature $f$. $S(i)$ denotes the entropy value of node $i$ (Gini index or any other split criterion may also be used), $n_i$ denotes the number of samples in node $i$ and $N$ denotes the overall number of samples in the training set. $l_i$ is the left child of node $i$ and $r_i$ is the right child. $E$, which is used to denote the weighted entropy value of a child node, is calculated as follows:
\begin{equation}
E(l_i)=S(l_i) \cdot \frac{n_{l_i}}{n_i}
\end{equation}
This importance calculation is similar to what is performed in the widely used \emph{scikit-learn} library\footnote{\url{https://scikit-learn.org/}}. Importance values are in the range of 0 to 1, and the sum of importance values over all features is 1.

We employed three different techniques to reduce the importance of a feature in tree-based models. These are described in the following sub-sections. Each method was first implemented in a simple decision tree. These special tree implementations were in turn used as a basis for implementing additional tree-based models, such as random forest and AdaBoost. In the ensemble-based methods, the limitations described were applied separately on each tree. 

\subsection{Weight-based Penalty}
\label{sub:weight}
In this method, each feature is assigned a weight according to its sensitivity level: the higher the sensitivity, the higher the weight. Weight values are in the range of 0 to 1, 0 signifying a non-sensitive feature. During the training process, when searching for the split with minimum entropy (or Gini index) at each stage, we use the feature's weight as a penalty. For a node $i$ that splits on feature $f$ with weight $w_f$, we replace the original entropy of the node with:
\begin{equation}
S(i) \cdot (1+w_f)
\end{equation}
(The addition of 1 ensures that a feature with weight 0 will remain with its original entropy). This is analogous to the class weighting method described by Polo et al. \cite{Polo07}. However, they assigned different weights to different classes in order to represent the relative importance of each class, whereas we assign different weights to different features.
After performing this weighted entropy calculation, features with higher sensitivity (i.e., weight) will yield higher entropy values, thus penalizing splits on more sensitive features.

\subsection{Limiting the Level of Splits}
\label{sub:level}
Building on an idea previously proposed as a countermeasure against tree inversion \cite{Fredrikson15}, this method limits the levels at which splits can be made on a sensitive feature. In the original method proposed by Fredrikson et al., the sensitive feature is only considered for splitting after $l-1$ other features have already been selected and removed from consideration afterwards. However, we define a threshold for the highest level in the tree at which the sensitive features may be split. In other words, at any level higher than the threshold, these features are not considered for splitting; once the threshold is reached, they may be considered for any subsequent split. We consider the level as the distance from the root of the tree. 

This method does not directly limit the number of splits that can be performed on a certain feature. That said, it does affect the feature's importance by forcing it to be used in splits that affect fewer samples (the higher the split in the tree, the more samples it affects, and vice versa).

The threshold level for being considered for a split is customizable, and may differ between datasets and model types. Intuitively, the deeper the trees in the initial model, the higher this threshold will have to be in order to effectively limit the importance of sensitive features. The results presented in Section \ref{sec:eval} were obtained by varying this parameter to yield different importance/accuracy trade-offs.

\subsection{Limiting the Number of Splits}
\label{sub:split}
Since the importance of a feature is summed over all nodes that split on that feature, we decided to try directly limiting the number of splits on sensitive features. The number of splits for each feature is counted. Once the threshold is met, that feature is no longer considered for splitting. To avoid creating an imbalance in the tree, where a certain feature is used heavily on the left side of the tree and then no longer used on the right side, we implemented a breadth-first-search (BFS) version of tree training. Thus, a feature is considered for splitting in a balanced manner near the top of the tree, and is no longer considered towards the bottom, once the maximum number of splits is reached.

The number of allowed splits is customizable, and may differ between datasets and model types. Here, again, the larger the initial trees (i.e., the number of internal nodes in the tree), the higher this threshold will have to be. The results presented in Section \ref{sec:eval} were obtained by varying this parameter to yield different importance/accuracy trade-offs.

We also implemented weighted variants of methods \ref{sub:level} and \ref{sub:split}, a normalized variant of method \ref{sub:split} (where the split number is normalized by the number of samples in each split), as well as different combinations of all of the above methods. However, these did not significantly improve our results.

\section{Evaluation}
\label{sec:eval}
We evaluated our method using two openly available datasets: the Nursery dataset\footnote{\url{https://archive.ics.uci.edu/ml/datasets/nursery}} and the GSS marital happiness dataset\footnote{\url{http://byuresearch.org/ssrp/downloads/GSShappiness.pdf}}. The Nursery dataset was derived from a hierarchical decision model originally developed to rank applications for nursery schools in Slovenia. It was used by Bild et al. \cite{Bild18} to compare approaches for differentially private statistical classification. The GSS marital happiness dataset is a subset of the GSS data created by Joseph Price to study various societal effects of pornography. It was used by Fredrikson et al. \cite{Fredrikson15} to evaluate the effectiveness of a model inversion attack. We removed records that did not have a value for the label feature, which was the case for almost half of the records in the GSS dataset. We also removed records whose label value was too scarce. This resulted in 12,958 samples in the Nursery dataset and 24,455 samples in the GSS dataset. Each dataset was divided into 80\% training and 20\% test.

We tested three types of tree-based models: Decision Tree (dt), Randofm Forest (rf), and AdaBoost (ab).

\subsection{Reducing Importance of Features}
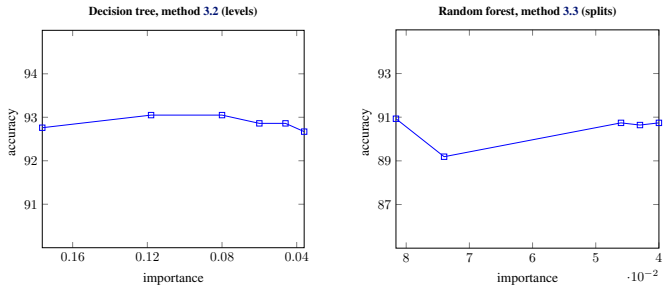
\begin{figure}
	\centering
	\hspace*{0.1in}
	\begin{subfigure}{.30\textwidth}	
	\resizebox{.70\columnwidth}{!}{%
	\begin{tikzpicture}[baseline,trim axis left]
	\begin{axis}[x tick label style={
		/pgf/number format/.cd,
		fixed,
		fixed zerofill,
		precision=2,
		/tikz/.cd
	}, legend style={font=\small}, title style={font=\small}, xlabel=importance, ylabel=accuracy, xmin=0.036,xmax=0.17626659, ymin=90, ymax=95,  x dir=reverse, ytick={91, 92, 93, 94}, xtick={0.04, 0.08, 0.12, 0.16},  title={\bfseries Decision tree, method \ref{sub:level} (levels)}, 
	legend pos=outer north east, at={(-1,0)},anchor=west, y label style={at={(0.07,0.4)},anchor=west}]
	\addplot [color=blue, mark=square] coordinates {( 0.036, 92.67 )(0.046,92.86)(0.06, 92.86)(0.08, 93.05)(0.118, 93.05)(0.17626659, 92.76)}; 
	\end{axis}
	\end{tikzpicture}}
\end{subfigure}%
\hspace*{-0.2in}
\begin{subfigure}{.30\textwidth}
	\resizebox{.70\columnwidth}{!}{%
		\begin{tikzpicture}[baseline,trim axis left]
		\begin{axis}[legend style={font=\small}, title style={font=\small}, xlabel=importance, ylabel=accuracy, xmin=0.04,xmax=0.08162115, ymin=85, ymax=95,  x dir=reverse, ytick={87, 89, 91,  93}, xtick={0.02, 0.03, 0.04, 0.05, 0.06, 0.07, 0.08},  title={\bfseries Random forest, method \ref{sub:split} (splits)}, 
		legend pos=outer north east, at={(-1,0)},anchor=west, y label style={at={(0.07,0.4)},anchor=west}]
		\addplot [color=blue, mark=square] coordinates {(0.04, 90.74)(0.043, 90.64)(0.046,90.74)(0.074, 89.19)(0.08162115, 90.93)}; 
		\end{axis}
		\end{tikzpicture}}
\end{subfigure}%
	\caption{Model accuracy vs. feature importance in Nursery data}
	\label{fig:nursery_graph}
\end{figure}

\begin{figure}
	\centering
	\hspace*{0.1in}
	\begin{subfigure}{.30\textwidth}
		\resizebox{.70\columnwidth}{!}{%
	\begin{tikzpicture}[baseline,trim axis left]
	\begin{axis}[x tick label style={
		/pgf/number format/.cd,
		fixed,
		fixed zerofill,
		precision=2,
		/tikz/.cd
	}, legend style={font=\small}, title style={font=\small}, xlabel=importance, ylabel=accuracy, xmin=0.031,xmax=0.16647047, ymin=61, ymax=66,  x dir=reverse, ytick={62, 63, 64, 65}, xtick={0.05, 0.1, 0.15},  title={\bfseries AdaBoost, method \ref{sub:weight} (weights)}, 
	legend pos=outer north east, at={(-1,0)},anchor=west, y label style={at={(0.07,0.4)},anchor=west}]
	\addplot [color=blue, mark=square] coordinates {(0.031	,63.36)(0.044, 62.95)(0.058, 63.05)(0.095, 63.61)(0.16647047, 63.1)}; 
	\end{axis}
	\end{tikzpicture}}
\end{subfigure}%
\hspace*{-0.2in}
\begin{subfigure}{.30\textwidth}
	\resizebox{.70\columnwidth}{!}{%
		\begin{tikzpicture}[baseline,trim axis left]
		\begin{axis}[legend style={font=\small}, title style={font=\small}, xlabel=importance, ylabel=accuracy, xmin=0,xmax=0.06263836, ymin=61, ymax=66,  x dir=reverse, ytick={62, 63, 64, 65}, xtick={0, 0.02, 0.04, 0.06},  title={\bfseries AdaBoost, method \ref{sub:level} (levels)}, 
		legend pos=outer north east, at={(-1,0)},anchor=west, y label style={at={(0.07,0.4)},anchor=west}]
		\addplot [color=blue, mark=square] coordinates {(0, 63.77)(0.003,63.51)(0.015,63.05)(0.029, 63.71)(0.06263836, 63.1)}; 
		\end{axis}
		\end{tikzpicture}}
	\end{subfigure}%
	\caption{Model accuracy vs. feature importance in GSS data}
	\label{fig:ab_gss_graph}
\end{figure}
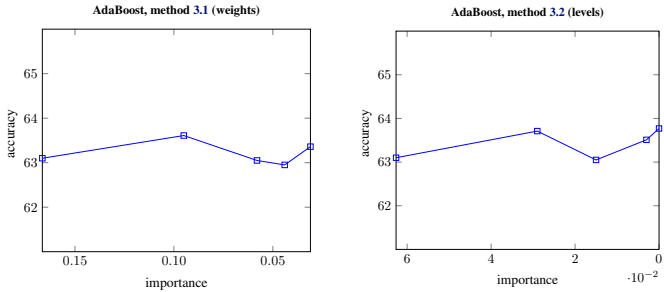

For each dataset, we first chose different random subsets of features to define as sensitive features. Next, we built a privacy-aware model that takes as input the sensitivity level of each feature. We recorded both the accuracy of the resulting model on our hold-out test dataset, as well as the new importance values of the selected features. 
We ran experiments on all combinations of dataset, model type and privacy-guided method (described in Section \ref{sec:priv}). Select results for a single feature are shown in Figures \ref{fig:nursery_graph} and \ref{fig:ab_gss_graph}. For reproducibility purposes, the threshold/weight values used in each run are brought in the Appendix.

Not all of the methods for reducing the importance of sensitive features worked well in all cases. Some methods worked better for certain model types, datasets, or features that we tested. Table \ref{table:methods} presents these results. We marked a combination of method and model type as successful (with an X) if we were able to find at least one feature for which the method worked. In many cases, we were able to reduce the importance without harming accuracy too much. This was even the case for some of the most important features (initially), as in the example of the `age' and `children' features in the GSS dataset, which were consistently found to be the most important features across all models before applying privacy-guided training. In some cases it was even possible to bring them down to 0 importance without harming accuracy, as shown in Figure \ref{fig:ab_gss_graph}. Surprisingly, in some cases, the accuracy was even improved. This is another indication that the choice of features to split made during model training is not always optimal.

This does not mean, however, that these mothods necessarily work for any feature in the dataset. In fact, there were some features whose importance could not be reduced without significantly harming the model's accuracy. 

\begin{table}
	\begin{tabular}{p{1cm}|p{1cm}||P{1.4cm}|P{1.3cm}|P{1.4cm}}
		Dataset & Method & Decision Tree & Random Forest & AdaBoost \\
		\hline \hline
		Nursery & Weights &  &  & X \\
		\hline
		 & Levels & X &  & \\
		\hline
		 & Splits & X & X & X \\
		\hline \hline
		GSS & Weights & X &  & X \\
		\hline
		 & Levels & X &  & X \\
		\hline
		 & Splits & X & X & X \\
	\end{tabular}
	\caption{Success of different privacy-guided methods}
	\label{table:methods}
\end{table}

We also conducted experiments with multiple sensitive features. These experiments were performed with sets of randomly selected features. Results for the Nursery dataset, decision tree model, and method \ref{sub:weight} (weights) are presented in Table \ref{table:num}. In addition to the model's accuracy, we show the sum of importance losses over all sensitive features. The higher this number, the better since this represents the degree by which the importance of all of these features decreased collectively. Our results show that even with relatively high numbers of sensitive attributes (up to 6 out of 14), it is possible to gain a meaningful reduction in the importance of these features with a very low impact on the model's accuracy. The last line in the table, displaying an accumulated importance loss of 0.264, corresponds to an overall initial importance of 0.507 summed over those 6 features; this thus represents a 50\% combined reduction in the importance of these features.

\begin{table}
	\begin{tabular}{p{2cm}||p{2cm}|p{2.8cm}}
		Num features & Model accuracy & Accumulated importance loss  \\
		\hline \hline
		1 & 92.57 &  0.044 \\
		\hline
		2 & 92.67 & 0.067 \\
		\hline
		3 & 91.22 & 0.117 \\
		\hline
		4 & 91.03 &  0.189 \\
		\hline
		5 & 91.61 &  0.236 \\
		\hline
		6 & 92.18 &  0.264 \\
	\end{tabular}
	\caption{Accumulated importance loss for multiple features}
	\label{table:num}
\end{table}

\subsection{Reducing Risk of Model Inversion}
Next, we chose one categorical feature in each dataset that had a relatively high importance, and could not simply be removed from the model without causing a significant decline in model accuracy. We then tried to infer the value of that feature using black-box (bb) and white-box (wb) model inversion attacks. The reasoning behind this choice was that if a feature can simply be removed without harming the accuracy of the model, there is no need for any special ``privacy-aware'' training process. It can simply be removed from the training. We found this to be the case, for example, for the `X-rated movies' feature that was used in the inversion attacks described by Fredrikson et al.  \cite{Fredrikson15}. 

For the GSS dataset we chose to attack the \emph{happiness} feature, which had an initial importance of 0.315. For the Nursery dataset we chose the \emph{social} feature, with an initial importance of 0.043. For simplicity of the attack implementation, we transformed each of these features to be Boolean, i.e., a feature that can only receive two values, 0 and 1. For the happiness feature, we assigned the value 1 to any record containing the initial value `Very happy', and 0 for the rest, yielding a prior distribution of [0.6, 0.4] for the values 0 and 1, respectively. For the social feature, we assigned the value 1 to any record containing the initial value `problematic' and 0 for the rest, yielding a prior distribution of [0.66, 0.34]. Here we also removed any rows missing values for the attacked feature. 

\begin{figure*}
	\hspace*{0.3in}
	\begin{subfigure}{.45\textwidth}
		\resizebox{.95\columnwidth}{!}{%
	\begin{tikzpicture}[baseline,trim axis left]
	\begin{axis}[legend style={font=\small}, title style={font=\small}, xlabel=importance, ylabel=attack accuracy, xmin=0.012,xmax=0.049, ymin=55, ymax=72,  x dir=reverse, ytick={58, 61, 64, 67, 70}, xtick={0.01, 0.02, 0.03, 0.04},  title={\bfseries Nursery data, social feature}, 
	legend pos=outer north east, at={(-1,0)},anchor=west, y label style={at={(0.07,0.4)},anchor=west}]
	\addplot [color=violet, mark=diamond] coordinates {(0.049, 68.8) 	( 0.046, 68.7 )	(  0.04, 67.2 )	( 0.025,67.4 )	(  0.02, 66.9 ) (  0.012, 66.9 )}; 
	\addlegendentry{$wb$}
	\addplot [color=blue, mark=square] coordinates {(0.049, 64.1) 	( 0.046, 63.6 )	( 0.04, 60.2 )	( 0.025, 59.5 )	( 0.02, 58.1 ) (  0.012, 58.5 )}; 
	\addlegendentry{$bb$}
	\addplot [color=black, mark=o] coordinates {(0.049, 58.2) 	( 0.046, 58.2 )	(  0.04, 58 )	( 0.025,58.6 )	(  0.02, 58.5 ) (  0.012,58.7 )};
	\addlegendentry{$ideal$}
	\end{axis}
	\end{tikzpicture}}
\end{subfigure}%
\hspace*{0.2in}
\begin{subfigure}{.45\textwidth}
	\resizebox{.95\columnwidth}{!}{%
	\begin{tikzpicture}[baseline,trim axis left]
	\begin{axis}[x tick label style={
		/pgf/number format/.cd,
		fixed,
		fixed zerofill,
		precision=2,
		/tikz/.cd
	}, legend style={font=\small}, title style={font=\small}, xlabel=importance, ylabel=attack accuracy, xmin=0.005,xmax=0.315, ymin=50, ymax=85,  x dir=reverse, ytick={55, 60, 65, 70, 75, 80}, xtick={0.005, 0.05, 0.1, 0.15, 0.2, 0.25, 0.3},  title={\bfseries GSS data, happiness feature}, 
	legend pos=outer north east, at={(-1,0)},anchor=west, y label style={at={(0.07,0.4)},anchor=west}]
	\addplot [color=violet, mark=diamond] coordinates {(0.315, 80) 	( 0.08, 74.6 )	(  0.061, 68 )	( 0.041,63.3 )	(  0.005, 59.9 )}; 
	\addlegendentry{$wb$}
	\addplot [color=blue, mark=square] coordinates {(0.315, 73.6) 	( 0.08, 69.1 )	( 0.061, 61.3 )	(0.041, 59.3 )	( 0.005, 56 ) }; 
	\addlegendentry{$bb$}
	\addplot [color=black, mark=o] coordinates {(0.315, 59.2) 	( 0.08, 58.8)	(  0.061, 58.9 )	( 0.041, 59.1 )	(  0.005, 58.6 ) };
	\addlegendentry{$ideal$}
	\end{axis}
	\end{tikzpicture}}
\end{subfigure}
	\caption{Attack accuracy vs. feature importance}
	\label{fig:attack_graph}
\end{figure*}
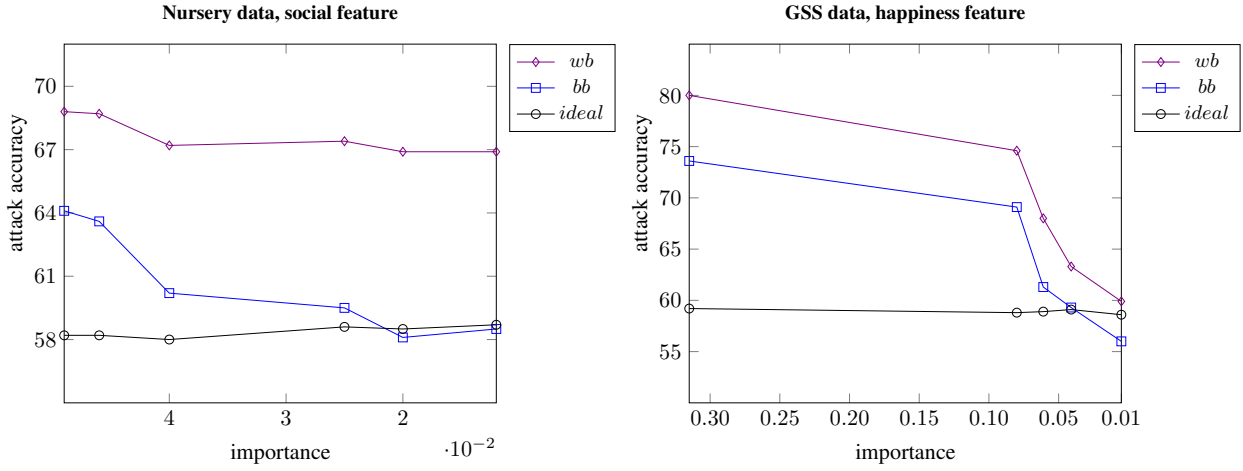

We recorded the accuracy of each type of attack, once using the original model and once using the equivalent privacy-aware model. The description of each of the attacks appears in the Appendix. We also compared these with the results of what we consider a ``baseline'' attack, referred to as the \emph{ideal adversary} in \cite{Fredrikson15}. This attack consists of training an ML model on the original dataset to predict the sensitive attribute. In our case, we used a neural network model for this ``ideal'' attack. Those results are presented in Figure \ref{fig:attack_graph}. 

These results demonstrate that it is indeed possible to reduce the success rate of model inversion attacks by reducing the importance of the sensitive (attacked) feature in the underlying model, as suggested by previous work. The baseline `ideal' attack, which does not depend on the model, is not directly affected by these changes and remains more or less constant, as we expected.

We performed two additional experiments with the aim of disproving the argument that the attacks' accuracy may deteriorate due to a decline in the original model's accuracy. In the first experiment, we used our privacy-guided training algorithm to generate models with decreased accuracy due to the reduced importance of other features (not the attacked feature). We show in Figure \ref{fig:static_graph} that the success rate of the attacks on the sensitive feature is not affected by this reduction in model accuracy and remains more or less static. This demonstrates that decreasing model accuracy alone, while keeping the attacked feature's importance at the same level, is not effective in countering attribute inference attacks.

In the second experiment, which was done on the GSS dataset, we were able to use our privacy-guided training algorithm to cause an \textbf{increase} in the attacked feature's importance (by decreasing the importance of other features in the model). In this experiment, we observed that the attacks' success increases with the increase of the feature's importance relative to the original model, with almost constant model accuracy. The results are shown in Figure \ref{fig:increase_graph}. As expected, the ideal attack is not affected by these changes as it does not depend on the model.

\begin{figure}
	\hspace*{0.3in}
	\resizebox{.88\columnwidth}{!}{%
	\begin{tikzpicture}[baseline,trim axis left]
	\begin{axis}[legend style={font=\small}, title style={font=\small}, xlabel=model accuracy, ylabel=attack accuracy, xmin=89.96,xmax=97.64, ymin=50, ymax=80, x dir=reverse, ytick={55, 60, 65, 70, 75}, title={\bfseries Attack accuracy vs. model accuracy in Nursery data},
	legend pos=outer north east, at={(-1,0)},anchor=west, y label style={at={(0.07,0.4)},anchor=west}]
	\addplot [color=violet, mark=diamond] coordinates {( 97.64, 69 )	( 96.48, 68.7 )	( 92.51, 68.9 )	( 90.97, 68.7 )	( 89.96, 66 )}; 
	\addlegendentry{$wb$}
	\addplot [color=blue, mark=square] coordinates {( 97.64, 63 )	( 96.48, 60.7 )	( 92.51, 61.6 )	( 90.97, 64.3 )	( 89.96, 63.5 ) }; 
	\addlegendentry{$bb$}
	\addplot [color=black, mark=o] coordinates {( 97.64, 58.2 )	( 96.48, 57.7 )	( 92.51, 59.1 )	( 90.97, 59.9 )	( 89.96, 58.9 )};
	\addlegendentry{$ideal$}
	\end{axis}
	\end{tikzpicture}}
	\caption{Decreasing accuracy of model}
	\label{fig:static_graph}
\end{figure}
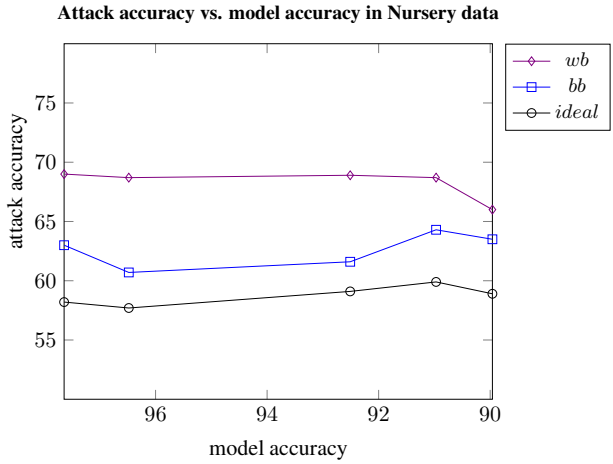

\begin{figure}
	\hspace*{0.3in}
	\resizebox{.88\columnwidth}{!}{%
	\begin{tikzpicture}[baseline,trim axis left]
	\begin{axis}[x tick label style={
		/pgf/number format/.cd,
		fixed,
		fixed zerofill,
		precision=1,
		/tikz/.cd
	}, legend style={font=\small}, title style={font=\small}, xlabel=importance, ylabel=attack accuracy, xmin=0.315,xmax=0.977, ymin=50, ymax=100, ytick={60, 70, 80, 90, 100}, xtick={0.3, 0.5, 0.7, 0.9},  title={\bfseries Attack accuracy vs. importance in GSS data}, 
	legend pos=outer north east, at={(-1,0)},anchor=west, y label style={at={(0.07,0.4)},anchor=west}]
	\addplot [color=violet, mark=diamond] coordinates {(0.315, 80) 	( 0.346, 79 )	(  0.504, 83.1 )	( 0.572,83.1 )	( 0.91, 99.6 )(  0.977, 99.5 )}; 
	\addlegendentry{$wb$}
	\addplot [color=blue, mark=square] coordinates {(0.315, 73.6) 	( 0.346, 74.5 )	( 0.504, 83 )	( 0.572, 83.1 )	( 0.91, 99.3 )( 0.977, 99.5 ) }; 
	\addlegendentry{$bb$}
	\addplot [color=black, mark=o] coordinates {(0.315, 59.2) 	( 0.346, 56.8 )	(  0.504, 59.5 )	( 0.572,57.3 )	( 0.91, 58.8 )(  0.977, 59.6 )};
	\addlegendentry{$ideal$}
	\end{axis}
	\end{tikzpicture}}
	\caption{Increasing importance of happiness feature}
	\label{fig:increase_graph}
\end{figure}
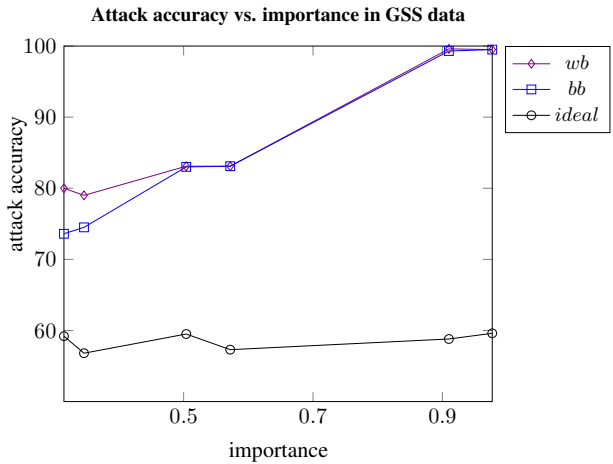

\section{Related Work}
\label{sec:rel}
\subsection{Membership Inference and Other Attacks}
There have been several attempts at \emph{membership inference attacks}, where one can deduce whether a specific individual was part of the training set or not. Examples include the work of Shokri et al. \cite{Shokri17} and Salem et al. \cite{Salem19}. As a result, a great body of prior work has focused on protecting the privacy of the training dataset used to train machine learning models from such attacks. Approaches that add noise during the training process have been proposed to protect people participating in the training set and counter membership attacks \cite{Zhang18}. The idea behind this approach is to reduce the effect of any single individual on the model's outcome. Some have even applied the principle of \emph{differential privacy} to the training process \cite{Abadi16, Holohan19}, which provides more robust privacy guarantees. Papernot et al. \cite{Papernot18} show how Private Aggregation of Teacher Ensembles (PATE) can scale to learning tasks with large numbers of output classes and uncurated, imbalanced, training data with errors. The PATE approach transfers to a ``student'' model the knowledge of an ensemble of ``teacher'' models, with intuitive privacy provided by training teachers on disjoint data and strong privacy guaranteed by noisy aggregation of teachers' answers. 

However, a recent survey of existing techniques based on differential privacy \cite{Jayaraman19} found that current mechanisms for differentially private machine learning rarely offer acceptable utility-privacy trade-offs for complex learning tasks. Settings that provide limited accuracy-loss provide little effective privacy, and settings that provide strong privacy result in useless models. Moreover, it was recently shown that differentially private training mechanisms have a disproportionate effect on the accuracy of the model for underrepresented or complex subgroups, leading to an increased bias towards these populations \cite{Bagdasaryan19}. 

Moreover, most of these methods are aimed at preventing membership inference attacks and not attribute inference attacks. Fredrikson et al. \cite{Fredrikson14} specifically evaluated the effectiveness of differential privacy to prevent their attack and reached the conclusion that for privacy budgets that are effective at preventing the attack, the utility of the model is lethally damaged, potentially exposing patients to increased risk of stroke, bleeding events, and mortality. 

Recently, a technique based on adversarial regularization was proposed to protect models against membership inference \cite{Nasr18}. The researchers introduced a mechanism that can train models with membership privacy, while ensuring indistinguishability between the predictions of a model on its training data and other data points (from the same distribution). They achieved this by minimizing the accuracy of the best black-box membership inference attack against the model, using an adversarial training algorithm that minimizes the prediction loss of the model as well as the maximum gain of the inference attacks.

Ateniese et al. \cite{Ateniese15} devised a different type of attack on machine learning models, which was aimed at stealing trade secrets. Another type of attack proposed by He, Zhang, and Lee \cite{He19} compromises the privacy of inference data in collaborative deep learning systems. In this setting, when a deep neural network and the corresponding inference task are split and distributed to different participants, they show that one malicious participant can accurately recover an arbitrary input fed into the system by another participant.

\subsection{Defenses Against Attribute Inference}
Zhao et al. \cite{Zhao19} present a theoretical framework for privacy-preservation under the attack of attribute inference. They propose a minimax optimization formulation to protect the given attribute and analyze its privacy guarantees against arbitrary adversaries. They also prove an information-theoretic lower bound to precisely characterize the fundamental trade-off between utility and privacy. They then proceed to analyze different existing privacy preservation algorithms to see if they uphold the lower bound in practice, and characterize their trade-off between privacy and utility.

Attriguard \cite{Jia18} uses evasion attacks from adversarial machine learning to defend against attribute inference attacks. However their setting is quite different from ours. They assume that the attacker uses a machine learning classifier to infer target private attributes based on public data. They further assume that the defender does not have access to this model or to the private attributes. Their solution is based on adding noise directly to users' public data. In our work, we assume that the defender is the owner of the machine learning model, and has access to both public and private attributes of the users used to train the model. Our goal is to prevent an attacker from inferring private attribute values from the public data \textbf{and model}.

Huang et al. \cite{Huang17} present a novel context-aware privacy framework called generative adversarial privacy (GAP). Under GAP, learning the privacy mechanism is formulated as a constrained minimax game between two players: a privatizer that sanitizes the dataset in a way that limits the risk of inference attacks, and an adversary that tries to infer the private variables from the sanitized dataset. This solution is intended for settings where a data holder would like to publish a dataset in a privacy preserving fashion; it is not geared specifically for building privacy-preserving ML models.

MLPrivacyGuard \cite{Alves19} is a countermeasure against black-box confidence-based model inversion attacks. It is based on adding controlled noise to the output of the confidence function, causing numerical approximation of gradient ascent to be unable to converge. This method was shown to be effective against attacks that apply gradient ascent on the confidence returned by the model. 
Whereas this method may be capable of countering a certain class of attacks, our method does not assume any specific attack. We demonstrate its effectiveness on several types of attacks, both black-box and white-box. 

\subsection{Privacy Protection of Tree-Based Models}
Additional work aimed specifically at tree-based models includes protocols for privately evaluating decision trees and random forests \cite{WuD16}. This work was carried out in a client-server setting, where the server has a decision tree (or random forest) model and the client holds an input to the model. They aimed at preserving the following security property: at the end of the protocol execution, the server would learn nothing about the input, and the client would learn nothing about the model. Also based on the concept of secure multiparty computation, Fritchman et al. \cite{Fritchman18} presented a first framework for privacy-preserving classification of tree ensembles with application in healthcare. Obviously, both of these works tackle a very different setting than ours, which is to protect the privacy of the training set against attribute inference.

Truex et al. \cite{Truex17} revisited the concepts and techniques of privacy-preserving decision tree learning. They reviewed and compared the privacy notions and properties of three orthogonal, yet complimentary, technical frameworks: randomization-based data obfuscation, differential privacy, and secure multiparty computation, analyzing their effectiveness. 

\section{Conclusions}
\label{sec:con}
We presented a simple solution for countering model inversion attacks in tree-based models. 
 We proposed three complimentary methods for reducing the importance of certain features, and showed that, in many cases, it is possible reduce a feature's importance without necessarily harming the model's accuracy. We presented results for three different types of models: decision trees, random forests and AdaBoost. 
 
 We also validated that a feature's importance indeed plays a pivotal role in how vulnerable it is to inversion, and that reducing a feature's importance may help protect it from such inversion attacks; this was demonstrated through several black-box and white-box attacks. Moreover, we disproved the potential argument that the attacks' chance of success may drop simply due to a decline in the original model's accuracy.
 
In the future, we plan to extend this work to additional types of models, focusing on neural networks, which have grown in popularity. We also plan to investigate additional risk factors and countermeasures for model inversion.

\bibliography{paper}
\bibliographystyle{icml2019}

\newpage 

\begin{appendices}
	\section{Description of attacks used for evaluation}
	\label{sec:attacks}
	We implemented three attacks aimed at inferring the value of a sensitive feature of an individual who participated in the training of a machine learning model M. We assumed the worse case with regard to the information available to the attacker in each case. In all three attacks we assumed that the attacker has access to all attribute values except for the sensitive attribute for the given individual, as well as access to enough labeled training data (including the value of the sensitive/attacked feature that we want to learn). We later refer to this dataset as the \emph{attack training set}. In all cases we used the attack training set to train an attack model, and then measured its accuracy on a test set.
	
	In the black-box attack we assumed that the attacker also has knowledge of the model's output (classification) for the given individual; and in the white-box attack we assumed that the attacker has knowledge of the model's internals, including the number of samples in the training set that were mapped to each leaf in the tree. This is a similar setup to the one described in (Fredrikson et al., 2015).
	Each of these attacks is described in more detail in the following subsections.
	
	\subsection{Baseline attack}
	In this attack we did not use the ML model at all, we only assumed access to all attribute values except the sensitive attribute for a given individual. In this attack we basically trained a simple neural network model on the none-sensitive attributes (on the attack training set) and tried to predict the value of the sensitive attribute (on the test set). In our implementation we used the default configuration of scikit-learn's MLPClassifier, i.e., one hidden layer with 100 neurons, relu activation and adam optimization.
	We use this as a baseline to compare to the other two model inversion attacks.
	
	\subsection{Black-box attack}
	This attack is almost the same as the baseline attack except that it uses the output (classification) of the model M in addition to the none-sensitive attribute values when training the neural network. Again we used the default configuration of scikit-learn's MLPClassifier to learn and predict the value of the sensitive attribute.
	
	\subsection{White-box attack}
	In this attack we assume full access to the model, i.e., the ability to use it to generate new predictions on inputs of our choice, as well as examine the internals of the model. We also use in this attack knowledge of the prior distribution of the attacked feature in the original training data, which we approximate by using the distribution in the attack training set. 
	
	We denote the vector of known feature values by $x_K$ and the known model prediction/classification for the full vector $x$ by $y$. We assume $x_1$ is the sensitive feature being attacked and denote by $v$ the specific values this feature may take.
	
	Given a sample $(x_K, y)$ to attack, we start by generating a dummy sample for each of the possible values for the sensitive (unknown) feature. In our case we used boolean features, so we created two dummy samples: $0,x_K$ and $1,x_K$. We then retrieved the model M's predictions for each of these dummy samples, denoted by $y_0$ and $y_1$ respectively. 
	
	If $y_0$ and $y_1$ are different and one of them matches the known value $y$, we use the corresponding value $v$ (0 or 1) as our prediction for the sensitive feature value. Otherwise, we use a simplified version of the calculation presented in (Fredrikson et al., 2015): 
	\begin{equation}
		P(x_1=v)=\sum_{i=1}^{m} p_i\phi_i(v) \cdot Pr[x_1=v] 
	\end{equation}
	We denote by the shorthand $P(x_1=v)$ the probability that $x_1=v$ given the auxiliary information. $\phi_i(v)$ is an indicator for the path (leaf) in the tree that corresponds to the sample $v,x_K$. When interating $i$ over all $m$ paths (or leaf nodes) in the tree, only one indicator receives the value 1 (the path that corresponds to the sample) and the rest receive the value 0. $p_i$ is the number of samples mapped to that leaf divided by the overall number of training samples ($n_i/N$) and $Pr[x_1=v]$ is the prior probability of the value $v$ in the attack training data. 
	
	We calculate the probability for each of the possible values $v$ (in our case 0 and 1) and choose the value with the highest probability.
	
	\section{Parameter Values Used in Evaluation}
	\label{sec:params}
	In this section we present the paramters used for each run presented in Figures 1-3 in the main paper. This information is brought in Tables \ref{table:dt_level}-\ref{table:gss3}; each table details the used parameter (weight, level number or split number, according to the chosen method), along with the corresponding importance value.
	
	\begin{table}
		\centering
		\begin{tabular}{p{2cm}|p{2cm}}
			Level & Importance  \\
			\hline \hline
			10 &  0.118 \\
			\hline
			25 & 0.08 \\
			\hline
			50 & 0.06 \\
			\hline
			75 &  0.046 \\
			\hline
			100 &  0.036 \\
		\end{tabular}
		\caption{Level values for Figure 1, Decision Tree model}
		\label{table:dt_level}
	\end{table}
	
	\begin{table}
		\centering
		\begin{tabular}{p{2cm}|p{2cm}}
			Splits & Importance  \\
			\hline \hline
			15 &  0.074 \\
			\hline
			12 & 0.043 \\
			\hline
			10 & 0.04 \\
			\hline
			9 &  0.046 \\		
		\end{tabular}
		\caption{Split values for Figure 1, Random Forest model}
		\label{table:rf_split}
	\end{table}
	
	\begin{table}
		\centering
		\begin{tabular}{p{2cm}|p{2cm}}
			Weight & Importance  \\
			\hline \hline
			0.2 &  0.095 \\
			\hline
			0.4 & 0.058 \\
			\hline
			0.6 & 0.044 \\
			\hline
			0.8 &  0.031 \\
			\hline
			1.0 &  0.032 \\
		\end{tabular}
		\caption{Weight values for Figure 2, Adaboost model}
		\label{table:ab_weight}
	\end{table}
	
	\begin{table}
		\centering
		\begin{tabular}{p{2cm}|p{2cm}}
			Level & Importance  \\
			\hline \hline
			400 &  0.029 \\
			\hline
			500 & 0.015 \\
			\hline
			750 & 0.003 \\
			\hline
			1000 &  0.0 \\
		\end{tabular}
		\caption{Level values for Figure 2, Adaboost model}
		\label{table:ab_level}
	\end{table}
	
	\begin{table}
		\centering
		\begin{tabular}{p{2cm}|p{2cm}}
			Splits & Importance  \\
			\hline \hline
			30 &  0.049 \\
			\hline
			10 & 0.046 \\
			\hline
			5 & 0.04 \\
			\hline
			3 &  0.025 \\
			\hline
			2 &  0.02 \\
			\hline
			1 &  0.012 \\
		\end{tabular}
		\caption{Split values for Figure 3, Nursery data}
		\label{table:nursery3}
	\end{table}
	
	\begin{table}
		\centering
		\begin{tabular}{p{2cm}|p{2cm}}
			Level & Importance  \\
			\hline \hline
			3000 & 0.08 \\
			\hline
			7000 & 0.061 \\
			\hline
			9000 &  0.041 \\
			\hline
			10000 &  0.005 \\
		\end{tabular}
		\caption{Level values for Figure 3, GSS data}
		\label{table:gss3}
	\end{table}
	
	Table \ref{table:nursery4} presents the number of splits versus the resulting model accuracy for the experiment whose results are brought in Figure 4 in the main paper. The following features were defined as sensitive when building the privacy-guided tree: children, health, has\_nurs, parents, housing and finance.
	
	\begin{table}
		\centering
		\begin{tabular}{p{2cm}|p{4cm}}
			Splits & Model Accuracy  \\
			\hline \hline
			50 &  97.64 \\
			\hline
			30 & 96.48 \\
			\hline
			10 & 92.51 \\
			\hline
			5 &  90.97 \\
			\hline
			3 &  89.96 \\
		\end{tabular}
		\caption{Split values for Figure 4}
		\label{table:nursery4}
	\end{table}
	
	Table \ref{table:gss5} presents the number of features defined as sensitive when building the privacy-guided tree versus the importance of the attacked feature (happiness) for the experiment whose results are brought in Figure 5 in the main paper.
	
	\begin{table}
		\centering
		\begin{tabular}{p{3cm}|p{4cm}}
			Num sensitive features & Importance of happiness feature  \\
			\hline \hline
			0 &  0.315 \\
			\hline
			9 & 0.346 \\
			\hline
			11 & 0.504 \\
			\hline
			14 &  0.572 \\
			\hline
			15 &  0.91 \\
			\hline
			17 & 0.977 \\
		\end{tabular}
		\caption{Split values for Figure 5}
		\label{table:gss5}
	\end{table}
	
\end{appendices}

\end{document}